\documentclass[final]{cvpr}

\usepackage{times}
\usepackage{epsfig}
\usepackage{graphicx}
\usepackage{amsmath}
\usepackage{amssymb}
\usepackage{multirow}
\usepackage[noend]{algpseudocode}
\usepackage{algorithmicx,algorithm}
\usepackage{rotating}
\usepackage[table]{xcolor}
\usepackage{setspace}

\usepackage[pagebackref=false,breaklinks=true,colorlinks,bookmarks=false]{hyperref}

\def\cvprPaperID{960} 
\def\confYear{CVPR 2021}

\begin{document}

\title{Delving into Data: Effectively Substitute Training for Black-box Attack}

\author{Wenxuan Wang$^{1*\ddagger}$\quad Bangjie Yin$^{2*\P}$\quad Taiping Yao$^{2*\P}$\quad Li Zhang$^{1\mathsection}$\quad \\ Yanwei Fu$^{1\dagger\mathsection}$\quad
Shouhong Ding$^{2\dagger\P}$\quad Jilin Li$^{2\P}$\quad Feiyue Huang$^{2\P}$\quad Xiangyang Xue$^{1\ddagger}$\\
\\
$^1$Fudan University \qquad $^2$Youtu Lab, Tencent}

\maketitle
\pagestyle{empty} 
\thispagestyle{empty} 

\renewcommand{\thefootnote}%
{\fnsymbol{footnote}}
\footnotetext[1]{indicates equal contributions.} 
\footnotetext[2]{indicates corresponding author.}
\footnotetext[3]{Wenxuan Wang and Xiangyang Xue are with School of Computer Science, and Shanghai Key Lab of Intelligent Information Processing, Fudan University. Email: \{wxwang19, xyxue\}@fudan.edu.cn.}
\footnotetext[4]{Li Zhang and Yanwei Fu are with School of Data Science, MOE Frontiers Center for Brain Science, and Shanghai Key Lab of Intelligent Information Processing, Fudan University. Email: \{yanweifu, lizhangfd\}@fudan.edu.cn.}
\footnotetext[5]{Bangjie Yin, Taiping Yao, Shouhong Ding, Jilin Li and Feiyue Huang are with Youtu Lab, Tencent. Email: \{bangjieyin, taipingyao, ericshding, jerolinli, garyhuang\}@tencent.com.}

\begin{abstract}
   Deep models have shown their vulnerability when processing adversarial samples. As for the black-box attack, without access to the architecture and weights of the attacked model, training a substitute model for adversarial attacks has attracted wide attention. Previous substitute training approaches focus on stealing the knowledge of the target model based on real training data or synthetic data, without exploring what kind of data can further improve the transferability between the substitute and target models.
   In this paper, we propose a novel perspective substitute training that focuses on designing the distribution of data used in the knowledge stealing process. 
   More specifically, a diverse data generation module is proposed to synthesize large-scale data with wide distribution. And adversarial substitute training strategy is introduced to focus on the data distributed near the decision boundary. The combination of these two modules can further boost the consistency of the substitute model and target model, which greatly improves the effectiveness of adversarial attack. 
   Extensive experiments demonstrate the efficacy of our method against state-of-the-art competitors under non-target and target attack settings. Detailed visualization and analysis are also provided to help understand the advantage of our method.
\end{abstract}

\section{Introduction}

Despite achieved impressive performance in most computer vision tasks, deep neural networks (DNNs) have been shown to be vulnerable to even imperceptible adversarial noise/perturbations~\cite{szegedy2013intriguing,luo2018towards}. The existence of adversarial examples reveals important security risks in deploying DNNs to real-world applications. The community studies the adversarial attacks in the settings of white-box or black-box attack, by whether or not fully access to the target attack model. Practically, as the information of the full target model for white-box attack is unavailable to real-world deployment, this paper particularly focuses on the black-box attack, which normally produces the adversarial examples only replying on hard-labels or output scores of the target model. Typically, the black-box attack includes the score-based \cite{chen2017zoo,ilyas2018prior,ilyas2018blackbox, dong2019efficient} or decision-based methods ~\cite{cheng2018queryeff, brendel2017decision}. Nevertheless, it is required to make an avalanche of queries to the target model in such attacks, still potentially limiting their usability to attack DNNs in real situations.

\begin{figure}[t]
\begin{center}
\hspace{-0.23in}
\includegraphics[scale=0.37]{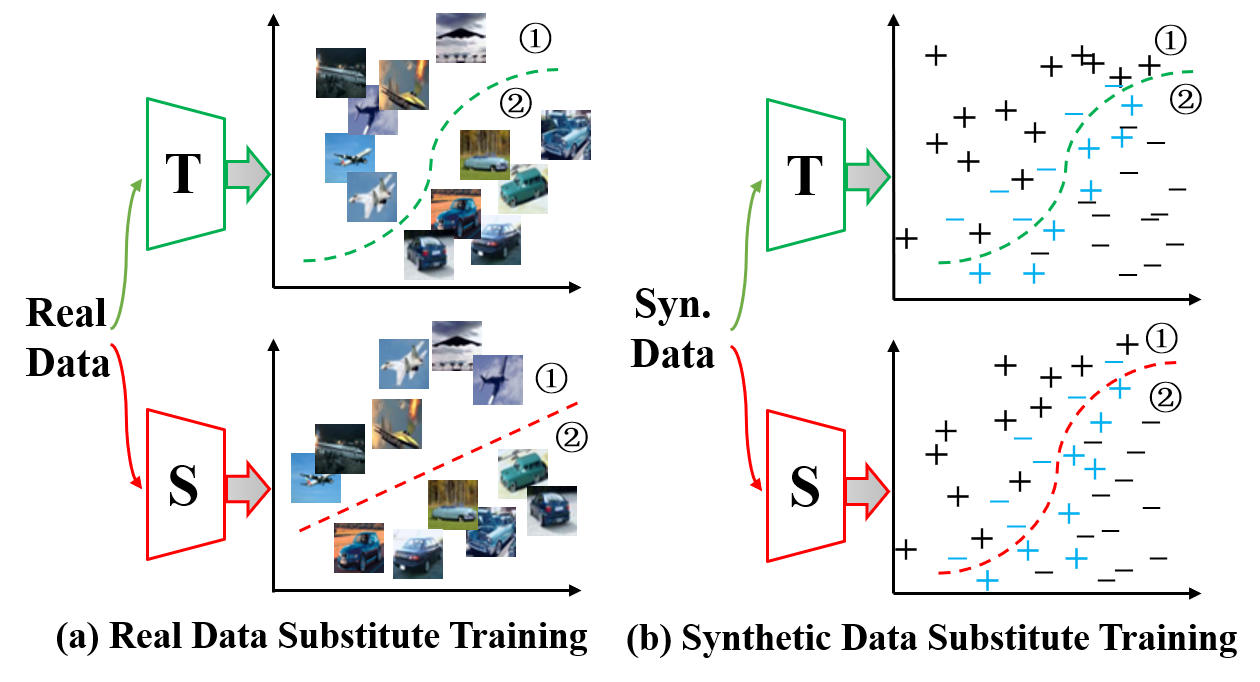}
\end{center}
   \caption{Differences between applying real data and synthetic data for substitute training. The `T'/`S' means the target/substitute model, the blue (+)/(-) in (b) indicates the adversarial examples, and the dashed green/red lines represent the decision boundary. Comparing (a) and (b), synthetic data generated in our way can train a substitute model with a more similar decision boundary to the target model. Best viewed in color and zoomed in.
   \label{fig:intro}}
\end{figure}

Recently, the idea of substitute training has been extensively explored in the black-box attack~\cite{goodfellow2014explaining, shi2019curls, kurakin2017adversarial, papernot2017practical, tramer2016stealing}. Normally, rather than directly learning to synthesize adversarial examples, a substitute model is trained to make similar predictions as the target model, queried by the same input data.  Within a certain amount of queries, this type of method is usually capable of learning a substitute model from the target model. Attack can thus be conducted on substitute model, and then transferable to the target model. 


Fundamentally,  substitute model tries to gain knowledge from the target model, by giving the input data and corresponding queried labels. Critically,
shall the input data come from the training data for the target model?
By assuming the `yes' answer, it indeed simplifies the substitute training.
However, it is even non-trivial to collect real input data in many real-world vision tasks. 
For example, the data of person images and videos are under very strict control, and the privacy of personal data has been well protected by the laws in many countries. 
Moreover, are the real images the most effective data for substitute training? The training data of the target model indeed help to get a well-performing substitute model on original task, but it cannot guarantee the transferability of the attack from the substitute model to target model, which has been proved in Tab.~\ref{Tab: Comparsion-P} and Tab.~\ref{Tab: Comparsion-L}.
For improving the attack performance in substitute training, it is necessary to minimize the decision boundary distance between the substitute and target models, which needs not only large-scale and diverse training data, but especially the data distributed near the decision boundary.

To address the limitation of real data and explore a better distribution of substitute training data, we propose a novel task-driven unified framework, which only uses specially-designed generated data for substitute training and achieves high attack performance. 
As shown in Fig.~\ref{fig:intro}, compared with using the training data of the target model to conduct substitute training, diverse synthetic data combined with adversarial examples will promote the substitute model to further approach the target.
More specifically, in our framework,
we first propose a novel Diverse Data Generation module (DDG), which samples noise combined with label-embedded information to generate diverse training data. Such distributed generated data can basically guarantee the substitute model to learn knowledge from the target.
Moreover, to further encourage the substitute model with similar decision boundary as the target,
Adversarial Substitute Training strategy (AST) is proposed to introduce adversarial examples as boundary data into the training process. Overall, the jointly learning of DDG and AST ensures the consistency between the substitute and target model, which greatly improves attack success rate in substitute training for black-box attack without any real data beforehand. 

The main contributions of this work are summarized as,
(1) We propose a novel effective generation-based substitute training paradigm to boost data-free black-box attacking performance, for the first time, by delving into the essence of input generated substitute training data.
(2) To achieve this goal, we firstly propose a diverse data generation module with multiple diverse constraints to broaden the distribution of synthetic data. 
And then further improve the consistency of decision boundaries between substitute model and target model by adversarial substitute training strategy. 
(3) The comprehensive experiments and visualizations over the four datasets and one online machine learning platform demonstrate the effectiveness of our method against the state-of-the-art attacks.

\section{Related work}
\noindent \textbf{Adversarial attack.}
Many previous works focus on the white-box attack ~\cite{szegedy2013intriguing, papernot2016limitations, carlini2017CW, kurakin2017adversarial, madry2017pgd} by generating adversarial examples through accessing gradient-information of target model. Furthermore, there are also some white-box attack methods studying the transferable attacking performance on the unknown black-box models ~\cite{dong2018boost, zhou2018transferable, chen2020attention}. Unfortunately, such a white-box setting greatly and unrealistically simplifies the attack task in the real-world scenario, as it demands a strong pre-condition of accessing the target models. In contrast, recent efforts are made on black-box attack methods, which has a more practical setting.  Normally, the attacker can only obtain the output scores or hard labels of a target victim model. 
In general, the black-box attack ~\cite{guo2019simple, brendel2017decision} is conducted by finding  adversarial examples from trials, which will cross the decision boundary of classes. For example, when processing the class probability output, Chen \textit{et al.} ~\cite{chen2017zoo} propose  utilizing a derivative of zeroth order to estimate the real gradients, and the work has been expanded by \cite{tu2019autozoom}. Ilyas \textit{et al.} ~\cite{ilyas2018blackbox, ilyas2018prior} also propose 
performing score-based black-box attack by prior knowledge. Nevertheless, previous black-box attacks are limited to prohibitive cost for extensively querying the target model, and significant number of real data for the corresponding target model. Rather than directly discovering the adversarial examples, our model learns to effectively synthesize the data distribution of target model for training a substitute model. Such a substitute model potentially saves plenty amount of queries to the target model during the attack generation. 
 
\begin{figure*}[htbp]
\begin{centering}
\includegraphics[scale=0.52]{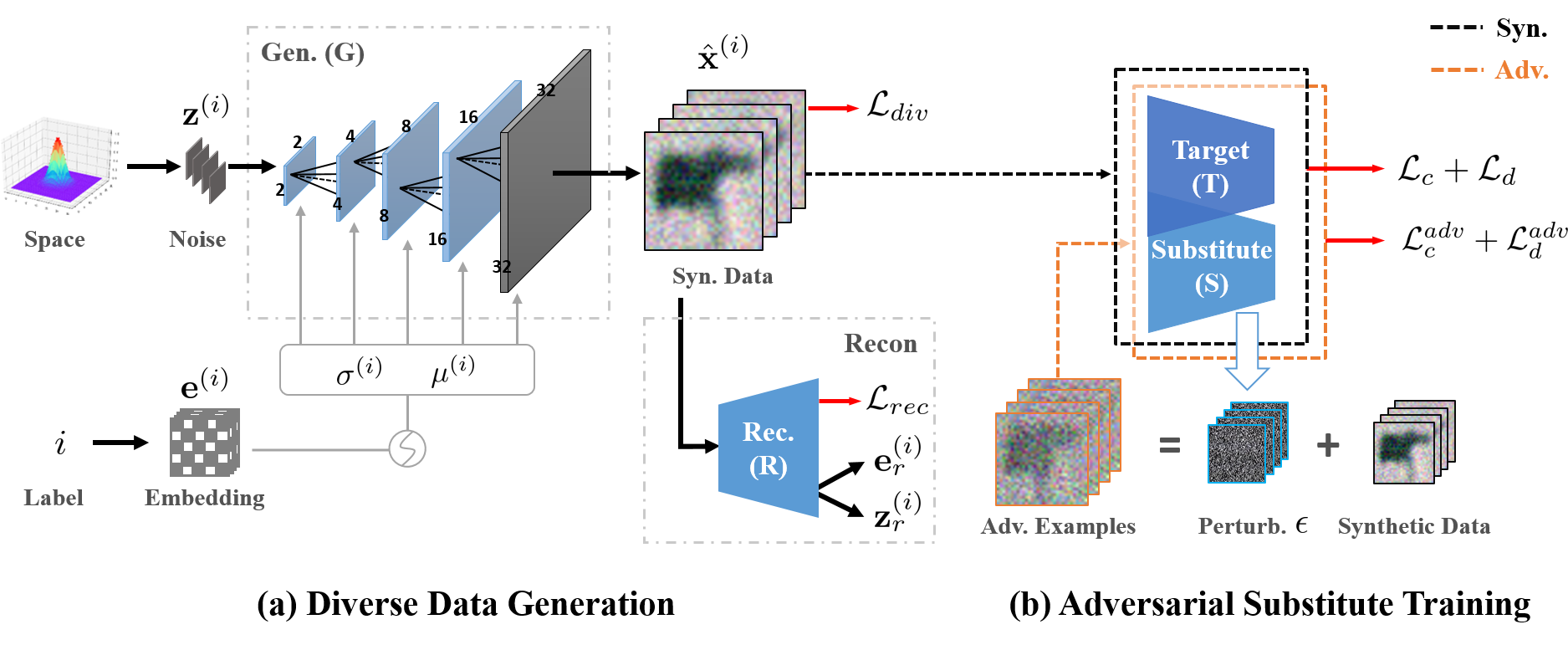} 
\par\end{centering}
\caption{Illustration of the unified proposed architecture, which consists of Diverse Data Generation module (DDG) and Adversarial Substitute Training module (AST). (a) DDG aims to generate diverse data with given label, which used to train a substitute model. (b) AST utilizes the adversarial examples generated from current substitute model to push the substitute model mimic the boundary of the target.
\label{fig:framework}}
\end{figure*}

\noindent \textbf{Substitute training.}
Substitute training is becoming a flourishing research direction. 
Papernot \textit{et al.} ~\cite{papernot2017practical} train the substitute model by utilizing a group of real images, and model theft attacks ~\cite{tramer2016stealing,zhu2020hermes} steal the target model also based on real data.
However, considering the privacy or unattainable problems of training data, some works ~\cite{yin2020dreaming, yoo2019knowledge, zhou2020dast} generate synthetic data to train a substitute model. 
Methods in ~\cite{yin2020dreaming, yoo2019knowledge} generate synthetic images from noise or recover training images from teacher model for substitute training based on
knowledge distilling (KD).
Zhou \textit{et al.} ~\cite{zhou2020dast} firstly propose an attack method 
to learn a substitute model under data-free condition. However, they only learn to output same results with target model, instead of further recovering the data distribution and decision boundary of the target, which are more crucial for the transferability of adversarial examples.
Different from their strategy, our proposed method starts from focusing on the distribution of generated data training for substitute model, comprehensively improving the attacking performance on black-box model by two perspectives of the diverse data generation and adversarial substitute training.

\section{Methodology}
\subsection{Framework Overview}
The objective of our work is to train a substitute model effectively for black-box adversarial attack, the whole proposed framework is illustrated in  Fig.~\ref{fig:framework}. 
It consists of two modules: Diverse Data Generation module (DDG) generating diverse data and Adversarial Substitute Training strategy (AST) further mimicking the `behaviour' of the target model.
In Fig.~\ref{fig:framework}(a), the DDG generates data $ \hat{x}^{(i)} = G(\textbf{z}^{(i)}, \textbf{e}^{(i)}) \ $ based on the random noise $ \textbf{z}^{(i)} $  and  label-embedded vector $ \textbf{e}^{(i)}$ for the label index $i $.
To guarantee the diversity of synthetic data, the generator $ G $ will be trained by three constraints, \textit{i.e.}, the adaptive label normalized generator, noise/label reconstruction, and inter-class diversity, which will be elaborated later.
Furthermore, to ensure the substitute model $S$ approximate 
the decision boundary of the target model $T$,
we feed the synthesized data along with the adversarial examples employed by AST into $S$ for substitute training in Fig.~\ref{fig:framework}(b). 
Essentially, we take the target model $T$ as a black-box of classifying $M$ classes, where only the label/probability outputs are available. The teacher-student strategy is re-purposed here to learn  $S$ from $T$. Finally, attacks  can  thus  be  conducted  on substitute model, and then transferable to the target model.

\subsection{Diverse Data Generation}
To synthesize better data for substitute training, we first propose a novel Diverse Data Generation module (DDG) with three constraints to manipulate the diversities of generated synthetic images. These constraints, in principle, encourage the generator $ G $ to learn relatively independent data-distribution for each different class, and keep the inter-class variances, which promote the alternative model to learn the knowledge of the target model.

\vspace{0.05in}
\noindent \textbf{Adaptive label normalized generator.}
To better learn from the target model, we need equally distributed data of all categories for substitute training, thus it is necessary to generate label-controlled data.
To realize that, we take full advantage of the given label and random noise.
Firstly, with the input of the random noise vector $ \textbf{z}^{(i)} \in \mathbb{R}^N $ sampled from standard Gaussian distribution and label \textbf{i}, we calculate the label-embedded vector $ \textbf{e}^{(i)} \in \mathbb{R}^N $ based on embedding layers \cite{mikolov2013distributed}.
Such label embedding process can encode a single discrete label to a continuous learnable vector, which has a wider distribution in the feature space and
contains more representation information. Unlike GANs, we have no real images for supervision, such a label embedding process is crucial for data generation.
Next, we extract the mean $ \mu^{(i)} $ and variance $ \sigma^{(i)} $ from the $ N $-dimensional label-embedding vector $ \textbf{e}^{(i)} $ by two full-connected layers. 
Then, the $ \mu^{(i)} $ and $ \sigma^{(i)} $ are involved in all deconvolution blocks to iteratively synthesize the image data with the condition of the specific category, which can be expressed as,
\begin{equation}
     \hat{\textbf{x}}_{t}^{(i)} = DeConv(\hat{\textbf{x}}_{t-1}^{(i)}) * \sigma^{(i)}  + \mu^{(i)} \label{x_update}
\end{equation}  
where there are total five de-convolution blocks, and $t$ represents the number of de-convolution block. After obtaining the final $ \hat{\textbf{x}}^{(i)} $, the output generated data has been decorated with label-normalized information. Such an adaptive label normalized generator can better leverage the relations between input noise and label-embedding vectors to synthesize label-controlled data.

\vspace{0.05in}
\noindent \textbf{Noise/Label reconstruction.}  
To further ensure the diversity of generated data $ \hat{\textbf{x}}^{(i)} $, 
we introduce a reconstruction net $R$ to reconstruct the input noise and label embedding $ \textbf{z}^{(i)}_r, \textbf{e}^{(i)}_r =  R(\hat{x}^{(i)}) $.  
And the corresponding reconstruction loss can be calculated as,


\begin{center}
\vspace{-8mm}
\begin{equation}
\mathcal{L}_{rec}=\sum_{i=0}^{M-1} \parallel (\textbf{z}^{(i)}_r - \textbf{z}^{(i)}) \parallel_1 + \mathrm{CE}(f(\textbf{e}^{(i)}_r,\textbf{e}), i)
\label{eq: reconstruction}
\end{equation}
\par\end{center}
where we use $ L_1 $ to denote the difference between the input $ \textbf{z}^{(i)} $ and reconstructed $ \textbf{z}^{(i)}_r $. 
As for label reconstruction, we apply function $f(*)$ to calculate the cosine distance between $ \textbf{e}^{(i)}_r $ and $ \textbf{e}$, which are further processed by Softmax to compute the cross entropy loss with the ground-truth label $ i $.
Under this constraint, our $ G $ can generate more diverse images for different input noise vectors of each class.

\vspace{0.05in}
\noindent \textbf{Inter-class diversity.}  To further enhance the data diversity of different classes, we 
use a cosine-similarity matrix to maximize the inter-class distance, for all the synthetic images. Particularly, the generator produces one input synthetic data batch of $ M_B \ll M $ different classes, and the model $S$ gives the output similarity matrix $ O_B \in \mathbb{R}^{M_B \times M_B}$ of this batch. Note that we have the ground-truth similarity matrix $ O_B^{gt} \in \left\{0, 1\right\}^{M_B \times M_B} $ with all the elements to be $ 0 $ except the diagonal elements are set to be $ 1 $. Thus the diversity loss function $ \mathcal{L}_{div} $ can be formulated as:
\begin{center}
\vspace{-4mm}
\begin{equation}
\mathcal{L}_{div}=\parallel TRI(O_B-O_B^{gt}) \parallel_2 
\label{eq: similarity}
\end{equation}
\par\end{center}
where $ TRI(*) $ is defined as an operation to extract the upper triangle elements of similarity matrix except the diagonal elements. In this way, $ \mathcal{L}_{div} $ will ensure the synthetic data owns the independent distribution for each class.

\subsection{Adversarial Substitute Training}

\begin{algorithm}[htb]
  \caption{The proposed black-box attack.}
  \label{alg:algorithm}
  \begin{algorithmic}
    \Require \\
      \hspace*{0.02in} {\bf Input:}
      Random noise $ \textbf{z}^{(i)} \in \mathbb{R}^N$;
      Label $ i \in \left\{0, 1, ...M-1\right\} $;
      Generator $ G $;
      Target victim model $ T $;
      Substitute model $ S $;
      Iterations $ R $.\\
      \hspace*{0.02in} {\bf Initialization:} 
      Model parameters $ \theta_G $, $ \theta_S $;
      hyper-parameters $\beta_1, \beta_2, \beta_3, \gamma_1, \gamma_2 $;.
  \end{algorithmic}
  \begin{algorithmic}[1]
    \Ensure
      Model parameters $ \theta_G^{*} $, $ \theta_S^{*} $.
    \State {\bf for} each $ r\in R $ {\bf do}
    \State \quad {\bf Synthetic data generation:} \\
     \qquad Given the label $i$ and random noise $\textbf{z}^{(i)}$, extract the mean $ \mu^{(i)} $ and variance $ \sigma^{(i)} $ from the label-embedded vector $ \textbf{e}^{(i)} $ \\
     \qquad Generate data through adaptive label normalized generator $\hat{\textbf{x}}_{t}^{(i)} = DeConv(\hat{\textbf{x}}_{t-1}^{(i)}) * \sigma^{(i)}  + \mu^{(i)}$ \\
     \qquad Generate adversarial examples based on the $\hat{\textbf{x}}^{(i)}$ 
     \State    \quad \textbf{Update S}: 
    \State \quad  Compute $ \mathcal{L}_S $ := ($ \mathcal{L}_{d} $, $ \mathcal{L}_{d}^{adv} $) then update $ \theta_{S}^{'} \leftarrow \theta_{S} - \gamma_1\bigtriangledown_{\theta_{S}} \mathcal{L}_{S}(\theta_{S}) $ 
    \State \quad \textbf{Update G}: 
     \State \qquad Calculate $ \mathcal{L}_G $ := ($ \mathcal{L}_{c} $, $ \mathcal{L}_{c}^{adv} $, $ \mathcal{L}_{rec} $, $ \mathcal{L}_{div} $) and then update $ \theta_{G}^{'} \leftarrow \theta_{G} - \gamma_2\bigtriangledown_{\theta_{G}} \mathcal{L}_{G}(\theta_{G}) $
    \State {\bf endfor}
    \State $ \theta_G^{*} = \theta_{G}^{'}$, $ \theta_S^{*} =  \theta_{S}^{'}$\\
    \Return $ \theta_G^{*} $, $ \theta_S^{*} $;
  \end{algorithmic}
\end{algorithm}

After DDG generates diverse training data,
for better attack performance, we still have to further encourage the substitute model with a more similar decision boundary as the target.
As is known to all, adversarial examples are wrongly classified 
with the visually-indistinguishable perturbations applied on. Due to the perturbations is relatively small, the adversarial examples can be seen as the samples around the decision boundary.
Therefore, we propose a novel adversarial substitute training strategy (AST), which utilizes the adversarial examples to further push the decision boundary of $ S $ more fitting to the $ T $'s. More specifically, for each iteration during training, our generator firstly synthesizes images through DDG. Then we choose the white-box attacking algorithm to obtain the adversarial perturbations $ \epsilon $ for the synthetic images based on the current $ S $. The objective function to generate adversarial images is defined as,
\begin{center}
\vspace{-8mm}
\begin{equation}
\min_{\epsilon \in [0, 1]^{d}}\ \parallel \epsilon \parallel + \lambda \cdot \mathcal{L}(\hat{x}^{(i)}+\epsilon, i^{adv})
\label{eq: attacks}
\end{equation}
\par\end{center}
where $ \mathcal{L}(\cdot) $ denotes an attack objective reflecting the probability or cross-entropy of predicting $ \hat{x}^{(i)}+\epsilon $ to be $ i^{adv} $, if considering the un-targeted attack, $ i^{adv} \neq i $, otherwise, $ i^{adv} = t $, $ t $ is a target label. $ \lambda $ is a regularization coefficient, and the constraint $ \epsilon \in [0, 1]^{d} $ confines the perturbation $ \epsilon $ to the valid image space. Then the generated images and corresponding adversarial data are used to updating $ S $ together.

\begin{table*}
\begin{spacing}{0.9}
\caption{Comparing ASRs results using probability as the target model output among our method and competitors over several datasets.
\label{Tab: Comparsion-P}}
\begin{centering}
\setlength{\tabcolsep}{1.8mm}{
\begin{tabular} {c}
\hspace{-0.1in}
\begin{tabular}{c|c|ccc|ccc|cc|cc}
\hline 
 & {\small{}Dataset } & \multicolumn{3}{c|}{{\small{}MNIST }} & \multicolumn{3}{c|}{{\small{}CIFAR-10 }} & \multicolumn{2}{c|}{{\small{}CIFAR-100 }} & \multicolumn{1}{c}{{\small{}Tiny ImageNet}}\tabularnewline
\hline 
 & {\small{}Target Model} & {\small{}AlexNet} & {\small{}VGG-16} & {\small{}ResNet-18} & {\small{}AlexNet} & {\small{}VGG-16} & {\small{}ResNet-18} & {\small{}VGG-19} & {\small{}ResNet-50} & {\small{}ResNet-50}\tabularnewline
\hline 
\hline 
\multirow{6}{*}{\begin{turn}{90}
{\small{}Non-Target}
\end{turn}} & {\small{}Training Data} & {\small{}41.36} & {\small{}29.25} & {\small{}34.81} & {\small{}30.95} & {\small{}23.15} & {\small{}32.66} & {\small{}14.47} & {\small{}18.33} & {\small{}12.86} \tabularnewline
 & {\small{}ImageNet} & {\small{}44.78} & {\small{}34.86}  & {\small{}31.39} & {\small{}36.84} & {\small{}22.94} & {\small{}34.01} & {\small{}17.26} & {\small{}20.93} & {\small{}21.75} \tabularnewline
\cline{2-11} 
 & {\small{}PBBA \cite{papernot2017practical}} & {\small{}52.53} & {\small{}50.31} & {\small{}59.77} & {\small{}45.82} & {\small{}30.19} & {\small{}33.91} & {\small{}22.34} & {\small{}28.11} & {\small{}26.54} \tabularnewline
 & {\small{}Knockoff \cite{orekondy2019knockoff}} & {\small{}59.21} & {\small{}58.38} & {\small{}65.82} & {\small{}50.93} & {\small{}31.58} & {\small{}39.40} & {\small{}27.73} & {\small{}29.55} & {\small{}29.99} \tabularnewline
 & {\small{}DaST \cite{zhou2020dast}} & {\small{}58.86} & {\small{}54.82} & {\small{}59.62} & {\small{}50.28} & {\small{}32.45} & {\small{}42.77} & {\small{}27.39} & {\small{}26.18} & {\small{}28.81} \tabularnewline
\cline{2-11}
 & \small{}Ours & \textbf{\small{}66.31} & \textbf{\small{}62.84} & \textbf{\small{}70.27} & \textbf{\small{}55.76} & \textbf{\small{}42.31} & \textbf{\small{}46.82} & \textbf{\small{}35.48} & \textbf{\small{}39.29} & \textbf{\small{}34.28} \tabularnewline
\hline 
\hline 
\multirow{6}{*}{\begin{turn}{90}
{\small{}Target}
\end{turn}} & {\small{}Training Data} & {\small{}38.45} & {\small{}40.27} & {\small{}43.94} & {\small{}11.45} & {\small{}10.35} & {\small{}11.22} & {\small{}5.02} & {\small{}8.66} & {\small{}6.17} \tabularnewline
 & {\small{}ImageNet} & {\small{}40.42} & {\small{}43.88} & {\small{}41.72} & {\small{}14.66} & {\small{}10.28} & {\small{}13.43} & {\small{}5.82} & {\small{}10.39} & {\small{}11.25} \tabularnewline
\cline{2-11}
 & {\small{}PBBA \cite{papernot2017practical}} & {\small{}42.67} & {\small{}55.66} & {\small{}49.24} & {\small{}25.83} & {\small{}15.38} & {\small{}20.44} & {\small{}6.73} & {\small{}17.22} & {\small{}13.88} \tabularnewline
 & {\small{}Knockoff \cite{orekondy2019knockoff}} & {\small{}48.28} & {\small{}52.89} & {\small{}54.27} & {\small{}30.87} & {\small{}16.92} & {\small{}19.56} & {\small{}12.83} & \textbf{\small{}22.37} & {\small{}15.26} \tabularnewline
 & {\small{}DaST \cite{zhou2020dast}} & {\small{}50.17} & {\small{}52.84} & {\small{}51.29} & {\small{}29.93} & {\small{}16.28} & {\small{}21.44} & {\small{}10.84} & {\small{}15.81} &  {\small{}13.92} \tabularnewline
\cline{2-11}
 & \small{}Ours & \textbf{\small{}59.29} & \textbf{\small{}57.28} & \textbf{\small{}64.46} & \textbf{\small{}33.81} & \textbf{\small{}29.89} & \textbf{\small{}25.77} & \textbf{\small{}17.23} & {\small{}21.44} & \textbf{\small{}19.37} \tabularnewline
\hline 
\end{tabular}
\end{tabular}}
\par\end{centering}
\end{spacing}
\end{table*}

\subsection{Loss Functions}

Finally, we apply the basic loss functions as in ~\cite{zhou2020dast} to train the  substitute model,

\begin{center}
\vspace{-4mm}
\begin{equation}
\mathcal{L}_{d} =\sum_{i=0}^{M-1}\parallel T(\hat{x}^{(i)}),S(\hat{x}^{(i)})\parallel_{F}
\label{eq: distance}
\end{equation}
\par\end{center}
\begin{center}
\vspace{-6mm}
\begin{equation}
\mathcal{L}_{c}  =e^{-\mathcal{L}_{d}}+\sum_{i=0}^{M-1}\mathrm{CE}(S(G(\textbf{z}^{(i)},\textbf{e}^{(i)})),i))
\label{eq: ce}
\end{equation}
\par\end{center}
\noindent where $ \mathcal{L}_d $ measures the distance between the output of $ T $ and $ S $, and $ \mathcal{L}_{c} $ denotes the generation loss. $ e^{-\mathcal{L}_d} $ implies a `min-max' game with $ \mathcal{L}_d $, $ \mathrm{CE}(\cdot) $ indicates the cross entropy loss between the prediction of $ S $ and input ground-truth label $ i $. Thus by virtue of such an alternately minimization of these two loss functions, the substitute model $ S $ can learn to mimic the outputs from the target model $ T $.
Further promoted by DDG and AST, with the generated data and adversarial examples, the unified substitute training loss $ \mathcal{L}_{S} $ and generator loss $ \mathcal{L}_{G} $  to train $S$ and $G$ are defined as,

\begin{center}
\vspace{-8mm}
\begin{equation}
\begin{split}
\mathcal{L}_{S}= \mathcal{L}_{d}+\mathcal{L}_{d}^{adv}
\end{split}
\label{eq: generator}
\end{equation}
\par\end{center}

\begin{center}
\vspace{-8mm}
\begin{equation}
\begin{split}
\mathcal{L}_{G}=
\beta_1 (\mathcal{L}_{c}+\mathcal{L}_{c}^{adv}) + \beta_2 \mathcal{L}_{rec} + \beta_3 \mathcal{L}_{div}
\end{split}
\label{eq: substitute}
\end{equation}
\par\end{center}
where $\mathcal{L}_{d}^{adv}$ is defined as the same as $\mathcal{L}_{d}$ in Eq. \ref{eq: distance} measuring the distance between the output from $T$ and $S$ using the adversarial examples as inputs.
$\mathcal{L}_{c}^{adv}$ is defined as $\mathcal{L}_{c}$ in Eq. \ref{eq: ce} to constrain the generation with adversarial examples as input, $\mathcal{L}_{rec}$ and $\mathcal{L}_{div}$ are proposed to enhance the diversity of generated data, , which will be elaborated later. $ \beta_1 $, $ \beta_2 $, and $ \beta_3 $ are the balanced hyper-parameters for DDG. Overall, the whole training process is illustrated in Alg.\ref{alg:algorithm}.


\begin{table*}
\begin{spacing}{0.9}
\caption{Comparing ASRs results using the label as target model output among our proposed method and competitors over several datasets. \label{Tab: Comparsion-L}}
\begin{centering}
\setlength{\tabcolsep}{1.8mm}{
\begin{tabular} {c}
\hspace{-0.1in}
\begin{tabular}{c|c|ccc|ccc|cc|cc}
\hline 
 & {\small{}Dataset } & \multicolumn{3}{c|}{{\small{}MNIST }} & \multicolumn{3}{c|}{{\small{}CIFAR-10 }} & \multicolumn{2}{c|}{{\small{}CIFAR-100 }} & \multicolumn{1}{c}{{\small{}Tiny ImageNet}}\tabularnewline
\hline 
 & {\small{}Target Model} & {\small{}AlexNet} & {\small{}VGG-16} & {\small{}ResNet-18} & {\small{}AlexNet} & {\small{}VGG-16} & {\small{}ResNet-18} & {\small{}VGG-19} & {\small{}ResNet-50} & {\small{}ResNet-50}\tabularnewline
\hline 
\hline 
\multirow{6}{*}{\begin{turn}{90}
{\small{}Non-Target}
\end{turn}} & {\small{}Training Data} & {\small{}17.45} & {\small{}20.11} & {\small{}24.50} & {\small{}13.76} & {\small{}10.43} & {\small{}13.05} & {\small{}5.01} & {\small{}8.58} & {\small{}7.32} \tabularnewline
 & {\small{}ImageNet} & {\small{}18.26} & {\small{}23.77} & {\small{}22.56} & {\small{}15.83} & {\small{}12.73} & {\small{}14.11} & {\small{}8.38} & {\small{}11.28} & {\small{}13.29} \tabularnewline
\cline{2-11}
 & {\small{}PBBA \cite{papernot2017practical}} & {\small{}22.45} & {\small{}28.18} & {\small{}29.00} & {\small{}21.84} & {\small{}13.63} & {\small{}17.66} & {\small{}11.48} & {\small{}16.33} & {\small{}15.37} \tabularnewline
 & {\small{}Knockoff \cite{orekondy2019knockoff}} & {\small{}25.39} & \textbf{\small{}33.18} & {\small{}37.72} & {\small{}20.16} & {\small{}20.74} & {\small{}19.87} & {\small{}16.48} & {\small{}18.31} & {\small{}22.33} \tabularnewline
 & {\small{}DaST \cite{zhou2020dast}} & {\small{}26.51} & {\small{}29.22} & {\small{}35.81} & {\small{}25.18} & {\small{}19.34} & {\small{}23.01} & {\small{}17.34} & {\small{}17.27} & {\small{}16.28} \tabularnewline
\cline{2-11}
 & \small{}Ours & \textbf{\small{}31.74} & {\small{}32.70} & \textbf{\small{}40.96} & \textbf{\small{}29.44} & \textbf{\small{}26.92} & \textbf{\small{}23.38} & \textbf{\small{}23.48} & \textbf{\small{}27.88} & \textbf{\small{}28.31} \tabularnewline
\hline 
\hline 
\multirow{6}{*}{\begin{turn}{90}
{\small{}Target}
\end{turn}} & {\small{}Training Data} & {\small{}15.53} & {\small{}12.55} & {\small{}10.88} & {\small{}9.92} & {\small{}10.24} & {\small{}9.09} & {\small{}3.97} & {\small{}6.44} & {\small{}4.92} \tabularnewline
 & {\small{}ImageNet} & {\small{}14.29} & {\small{}14.81} & {\small{}15.70} & {\small{}11.01} & {\small{}12.22} & {\small{}9.32} & {\small{}4.82} & {\small{}8.56} & {\small{}7.02} \tabularnewline
\cline{2-11}
 & {\small{}PBBA \cite{papernot2017practical}} & {\small{}15.26} & {\small{}19.86} & {\small{}18.53} & {\small{}12.84} & {\small{}11.33} & {\small{}10.48} & {\small{}6.91} & {\small{}7.33} & {\small{}8.61} \tabularnewline
 & {\small{}Knockoff \cite{orekondy2019knockoff}} & {\small{}19.48} & {\small{}23.74} & {\small{}17.85} & {\small{}16.38} & {\small{}12.80} & {\small{}13.91} & {\small{}9.48} & {\small{}9.52} & {\small{}10.65} \tabularnewline
 & {\small{}DaST \cite{zhou2020dast}} & {\small{}20.03} & {\small{}21.48} & {\small{}19.33} & {\small{}15.72} & {\small{}15.92} & {\small{}14.83} & {\small{}7.48} & {\small{}10.39} &  {\small{}10.31} \tabularnewline
\cline{2-11}
 & \small{}Ours & \textbf{\small{}25.56} & \textbf{\small{}27.64} & \textbf{\small{}21.83} & \textbf{\small{}21.66} & \textbf{\small{}18.67} & \textbf{\small{}17.90} & \textbf{\small{}12.47} & \textbf{\small{}16.26} & \textbf{\small{}13.39} \tabularnewline
\hline 
\end{tabular}
\end{tabular}}
\par\end{centering}
\end{spacing}
\end{table*}

\section{Experiments}
\subsection{Experiment Setup}
\noindent \textbf{Datasets and target model.}
1) MNIST \cite{lecun1998gradient}:
The attacked model is pre-trained on AlexNet \cite{krizhevsky2017imagenet}, VGG-16 \cite{simonyan2014very}, and ResNet-18 \cite{he2016deep}.
The default substitute model is a network with 3 convolutional layers.
2) CIFAR-10 \cite{krizhevsky2009learning}: 
The attacked is pre-trained on AlexNet, VGG-16, and ResNet-18.
The default substitute model is VGG-13.
3) CIFAR-100 \cite{krizhevsky2009learning}:
The attacked is pre-trained on VGG-19 and ResNet-50.
The default substitute model is ResNet-18. 
4) Tiny Imagenet \cite{russakovsky2015imagenet}:
The attacked is pre-trained on ResNet-50. 
The substitute model is ResNet-34.

\noindent \textbf{Competitors.} To verify the efficacy of the proposed method, we compare our attacking results with data-free black-box attack, \textit{i.e.}, DaST \cite{zhou2020dast}, and several black-box attacks which require real data, such as PBBA \cite{papernot2017practical} and Knockoff \cite{orekondy2019knockoff}. We also conduct substitute training using the original training data of the attacked model, and utilize ImageNet \cite{russakovsky2015imagenet} to learn the substitute model.

\begin{table}
\caption{Comparing ASRs results among our proposed method and competitors for attacking the Microsoft Azure example model. \label{Tab: Comparsion-Azure}}
\begin{centering}
\begin{tabular} {c}
\hspace{-0.1in}
\begin{tabular}{c|c|cc}
\hline 
 & Method & Probability-based & Lable-based\tabularnewline
\hline 
\multirow{4}{*}{\begin{turn}{90}
{\small{}Non-Target}\end{turn}} 
& PBBA \cite{papernot2017practical} & 82.34 & 80.29 \tabularnewline
 & Knockoff \cite{orekondy2019knockoff} & 88.91 & 92.88 \tabularnewline
 & DaST \cite{zhou2020dast} & 90.63 & 96.97 \tabularnewline
\cline{2-4}
 & Ours & \textbf{96.73} & \textbf{98.91} \tabularnewline
\hline 
\multirow{4}{*}{\begin{turn}{90}
{\small{}Target}\end{turn}} 
& PBBA \cite{papernot2017practical} & 39.23 & 49.39\tabularnewline
 & Knockoff \cite{orekondy2019knockoff} & 46.97 & 63.99\tabularnewline
 & DaST \cite{zhou2020dast} & 45.66 & 65.91\tabularnewline
\cline{2-4}
 & Ours & \textbf{57.92} & \textbf{69.81} \tabularnewline
\hline 
\end{tabular}
\end{tabular}
\par\end{centering}
\end{table}

\noindent \textbf{Implementation details.}
We use Pytorch for Implementation. We utilize Adam to train our substitute model, generator, and reconstruction net from scratch, and all weights are randomly initialized using a truncated normal distribution with std of 0.02. The initial learning rates of all networks are set as 0.0001, they are gradually decreased to zero from the 80th epoch, and stopped after the 150th epoch. We set the mini-batch size as 500, the hyper-parameters $ \beta_1, \beta_2$, and $ \beta_3 $ are equally as 1. Our model is trained by one NVIDIA GeForce GTX 1080Ti GPU. 
We apply PGD \cite{madry2017pgd} as the default method to generate adversarial images during the AST and evaluation. We also utilize FGSM \cite{goodfellow2014explaining}, BIM \cite{kurakin2016adversarial} and C\&W \cite{carlini2017CW} to conduct attack for extensive experiments. 

\noindent \textbf{Evaluation metrics.}
Considering there exist two different scenarios as proposed in DaST \cite{zhou2020dast}, \textit{i.e.}, only get the output label from the target model and access the output probability well, and we name these two scenarios as Probability-based and Label-based. In the experiments, we report the attack success rates (ASRs) of the adversarial examples generated by the substitute model to attack the target black-box model. Following the setting in DaST \cite{zhou2020dast}, in the non-target attack setting, we only generate adversarial examples on the images classified correctly by the attacked model. For target attacks, we only generate adversarial examples on the images which are not classified to the specific wrong labels. For a fair comparison, during all the adversarial example generation, we restrict the perturbation $\epsilon =8 $. We conduct five times over each testing, and report the average results.

\subsection{Black-box Attack Results}
We evaluate our method with competitors over four datasets and one online machine learning platform for both target and non-target attack settings. As shown in Tab.~\ref{Tab: Comparsion-P}, Tab.~\ref{Tab: Comparsion-L}, and Tab.~\ref{Tab: Comparsion-Azure}, we conduct extensive comparison with multiple target models for each dataset under both probability-based and label-based scenarios. 

\noindent \textbf{Comparisons with the real data for substitute training.} Here we study the substitute training for attacking with real images, as listed in Tab.~\ref{Tab: Comparsion-P} and Tab.~\ref{Tab: Comparsion-L}, we directly use the original training data of the target model or ImageNet to apply substitute training instead of the synthesized. The results show that the real image can let the substitute model learn a little from the target and may higher accuracy on the classification, but weaker attack strength compared to the generated data. We believe that this problem is caused by the number and diversity limitation of the real images, which may lead to failure of the substitute model learning and mimicking from the target one. Thus, we propose a DDG strategy to synthesize large-scale and diversified data.

\noindent \textbf{Comparisons with the state-of-the-art.} As shown in Tab.~\ref{Tab: Comparsion-P} and Tab.~\ref{Tab: Comparsion-L}, we compare our method with black-box attacks. 
For both non-target and target attacking settings, our method achieves the best ASRs over Probability-based and Label-based scenarios under all datasets. In addition, compared to similar generative DaST, our method significantly outperforms it with a large margin. The results verify the efficacy of proposed method to encourage the substitute model better approximate the target’s decision boundary and achieve high ASRs for data-free black-box attack. 

\noindent \textbf{Comparisons with competitors on Microsoft Azure.} To better evaluate the attack method ability under the real-world applications, we conduct experiments for attacking the online model on Microsoft Azure. Target at attacking the example MNIST model of the machine learning tutorial on Azure, we compare the results between our methods and competitors. The results shown in Tab.~\ref{Tab: Comparsion-Azure}, indicate our method can achieve the best ASRs over the online model, which further prove the efficacy of our method under the real scenario without prior knowledge of the attack one. 

\begin{table}
\caption{ASRs results of variants of the proposed attack method. The components are overlaid gradually with the rows. The target model is based on AlexNet for MNIST, and VGG-16 for CIFAR-10, the substitute models are the default ones according to the datasets. `C-100' refers to the CIFAR-100 dataset. \label{Tab: Abaltion}}
\begin{centering}
\setlength{\tabcolsep}{1.4mm}{
\begin{tabular} {c}
\hspace{-0.1in}
\begin{tabular}{c|c|cc|cc}
\hline 
\multirow{2}{*}{} & \multirow{2}{*}{Components} & \multicolumn{2}{c|}{Probability-based} & \multicolumn{2}{c}{Label-based}\tabularnewline
\cline{3-6}
 &  & MNIST & C-100 & MNIST & C-100\tabularnewline
\hline 
\multirow{6}{*}{\begin{turn}{90}
Non-Target \end{turn}}
& Baseline & 29.42 & 8.27  & 13.84 & 4.27 \tabularnewline
 & + ALNG & 49.18 &  21.38 & 20.85 & 12.66 \tabularnewline
 & + N/LR & 55.21 & 26.31 & 24.91 & 15.99 \tabularnewline
 & + ICR & 62.82 & 31.27 & 28.20  & 19.94 \tabularnewline
& + AST (Ours) & \textbf {66.31} & \textbf {35.48} & \textbf {31.74} & \textbf {23.48} \tabularnewline
\hline 
\hline 
\multirow{6}{*}{\begin{turn}{90}
Target \end{turn}}
& Baseline & 26.29 & 3.27 & 11.48 & 1.29 \tabularnewline
 & + ALNG & 44.48 & 10.47  & 16.28 & 7.83 \tabularnewline
 & + N/LR & 51.87 & 11.83 & 19.59 & 9.38 \tabularnewline
 & + ICR & 54.01 & 14.89 & 22.48 & 11.03 \tabularnewline
 & + AST (Ours) & \textbf {59.29} & \textbf {17.23} & \textbf {25.56} & \textbf {12.47} \tabularnewline
\hline 
\end{tabular}
\end{tabular}}
\par\end{centering}
\end{table}

\subsection{Ablation Study}
\subsubsection{Quantitative Results}
\noindent \textbf{The efficacy of different components in the proposed method.} To generate label-controlled and diversity data for substitute training and make the substitute model better fit the decision boundary of the target, our method applies the following components: 
(a) `Adaptive label normalized generator' (ALNG): generates data from input random noise with label-embedded vector; (b) `Noise/Label reconstruction' (N/LR): applies a reconstruction net to reconstruct the input noise and given label; (c) `Inter-class diversity' (ICR): constrains the distance of generated data in inter-classes; (d) `Adversarial substitute training' (AST): uses the adversarial examples to further train the substitute model. In Tab.~\ref{Tab: Abaltion}, we list the variants by adding the above components gradually, and the `Baseline' refers to the framework using random noise and given label as the contacted input to directly generate data and train substitute model.

As the results shown in the Tab.~\ref{Tab: Abaltion}, we can find that without ALNG, the substitute model `Baseline' can hardly learn knowledge from the attacked one, which may due to the poor generation without powerful controlled label constrains. Besides, models with the N/LR and ICR can achieve much higher ASRs results compared to the former, these verify that more diverse label-controlled generated data can make the substitute models learn more knowledge from the target. To ensure that the substitute model approximates the attacked decision boundary, the AST technique is applied to generate adversarial examples as the boundary data to make the substitute model mimic the attacked one and further boost the attack results. The ASRs results clearly demonstrate the important roles of components for data-free black-box attack.

\begin{table}
\caption{ASRs results applying various attacks to generate adversarial examples for AST under different attack evaluation in MNIST. The target model is AlexNet and the substitute model is the default. The 3rd and 4th columns represent applying FGSM to conduct AST, the last two columns utilize PGD for AST, and the raw indicates the attack to evaluate. The `-P' and `-L' in table mean the Probability-based and Label-based scenarios, respectively. \label{Tab: Attacks}}
\begin{centering}
\setlength{\tabcolsep}{2.6mm}{
\begin{tabular}{c|c|cc|cc}
\hline 
\multirow{2}{*}{} & \multirow{2}{*}{Attacks} & \multicolumn{2}{c|}{FGSM \cite{goodfellow2014explaining}} & \multicolumn{2}{c}{PGD \cite{madry2017pgd}}\tabularnewline
\cline{3-6}
 &  & -P & -L & -P & -L \tabularnewline
\hline 
\multirow{4}{*}{\begin{turn}{90}
Non-Target \end{turn}} & FGSM \cite{goodfellow2014explaining} & \textbf{70.26} & 36.29 & 57.35 & \textbf{33.10} \tabularnewline
 & BIM \cite{kurakin2016adversarial} & 66.38 & \textbf{36.97} & \textbf{68.45} & 29.58\tabularnewline
 & PGD \cite{madry2017pgd} & 62.63 & 33.72 & 66.31 & 31.74 \tabularnewline
 & C\&W \cite{carlini2017CW} & 49.92 & 20.91 & 46.93 & 22.02\tabularnewline
\hline 
\hline 
\multirow{4}{*}{\begin{turn}{90}
Target \end{turn}} & FGSM \cite{goodfellow2014explaining} & 50.82 & 27.38 & 29.48 & 19.25\tabularnewline
& BIM \cite{kurakin2016adversarial} & \textbf{67.29} & 32.33 & \textbf{44.82} & 18.14\tabularnewline
& PGD \cite{madry2017pgd} & 52.77 & \textbf{33.39} & 39.29 & \textbf{25.56}\tabularnewline
& C\&W \cite{carlini2017CW} & 49.38 & 20.39 & 28.57 & 19.66\tabularnewline
\hline 
\end{tabular}}
\par\end{centering}
\end{table}

\noindent \textbf{The affect of different attacks.} Consider that we require adversarial examples during the substitute training, here we evaluate the affect of attack method for our algorithm. As shown in Tab.~\ref{Tab: Attacks}, the column means the attacks to generate adversarial examples for substitute training, and the row refers to the method to attack the target model in evaluation. The results indicate that different attacks may not have obvious impacts on our method, which means that using different attacks for substitute training hardly affects the final attacking results. Thus, our method is effective under various attacks and there is no need to restrict the attack method between the substitute training and evaluation.

\begin{table}
\caption{ASRs results with different substitute models attacking VGG-16 trained on CIFAR-10. The `-P' and `-L' in table mean the Probability-based and Label-based scenarios, respectively.  \label{Tab: Structures}}
\begin{centering}
\begin{tabular}{c|cc|cc}
\hline 
 & \multicolumn{2}{c|}{Non-Target Attack} & \multicolumn{2}{c}{Target Attack}\tabularnewline
\hline 
 & -P & -L & -P & -L\tabularnewline
\hline 
AlexNet \cite{krizhevsky2017imagenet} & 39.78 & 22.57 & 24.90 & 18.45 \tabularnewline
VGG-13 \cite{simonyan2014very} & 42.31 & 26.92 & 29.89 & 18.67 \tabularnewline
VGG-16 \cite{simonyan2014very} & 45.24 & 25.28 & 30.41 & \textbf{22.46} \tabularnewline
VGG-19 \cite{simonyan2014very} & 45.92 & 27.69 & 32.59 & 21.94 \tabularnewline
ResNet-18 \cite{he2016deep} & \textbf{49.28} & 26.83 & \textbf{33.20} & 20.58 \tabularnewline
ResNet-34 \cite{he2016deep} & 48.94 & \textbf{28.72} & 30.48 & 20.4 \tabularnewline
\hline
\end{tabular}
\par\end{centering}
\end{table}

\noindent \textbf{The impact of different substitute model architecture.} We aim to achieve a successful attack for black-box under data-free condition, thus we have no prior knowledge of the attacked model structure. To further evaluate the impact of different substitute model architecture, we apply several substitute models for the same attacked one, which is VGG-16 pre-trained on CIFAR-10. As shown in Tab.~\ref{Tab: Structures}, we try various architecture as the substitute model, i.e., AlexNet, VGGNet, and ResNet, and the results show that there does not exist the most suitable structure which can achieve the best ASRs under all settings. Except for the simplest AlexNet, the others reach similar high ASRs results, which demonstrates that the different substitute model architecture may not have a huge impact on the attack strength, but still recommend choosing a deeper network.

\begin{figure}
\begin{centering}
\hspace{-0.2in}
\includegraphics[scale=0.22]{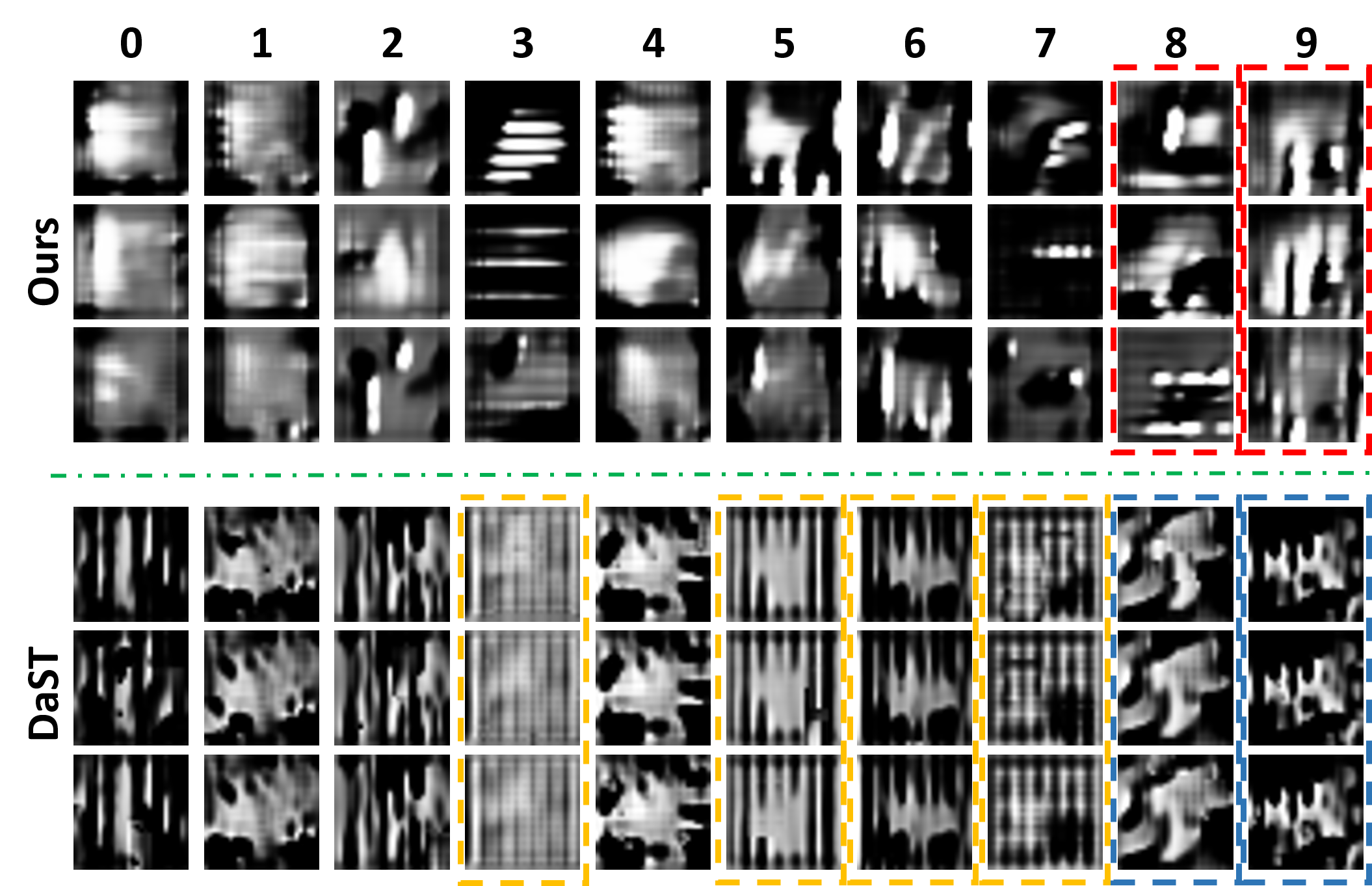} 
\par\end{centering}
\caption{Samples of generated images on MNIST. The upper part represents the images generated from ours, and the below from DaST. Left to right represents 0--9 hand-written number classes, and the samples belong to the same column have the same label.
\label{fig:generated images}}
\end{figure}

\begin{figure}
\begin{centering}
\includegraphics[scale=0.37]{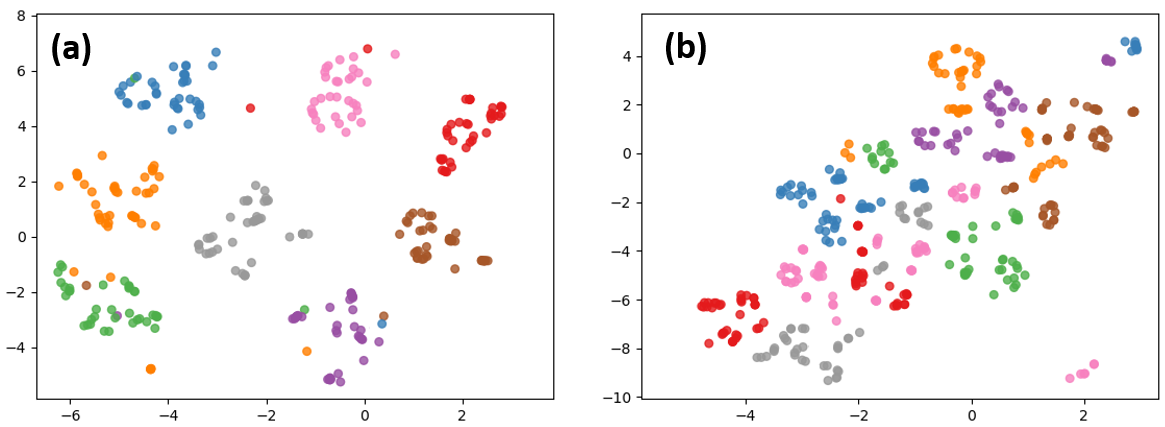} 
\par\end{centering}
\caption{Visualization of generated data in 8 classes (one color indicates one category) using t-SNE \cite{maaten2008visualizing} on CIFAR-100. (a) Data generated by DDG module. (b) Data generated by the DaST. 
\label{fig:tsne}}
\end{figure}

\subsubsection{Qualitative Results}
\noindent \textbf{Our model can improve the diversity of generated data across different categories.}
(1) As for the generated data shown in Fig.~\ref{fig:generated images}, we illustrate the data synthesized in different ways. It is clear that, compared with ours, the generated data of DaST without ALNG, N/LR, and ICD strategies, are much more similar in extra-classes, such as the data in yellow dashed boxes have the approximate vertical lines.
(2) In terms of feature, we visualize the feature distribution of synthesized data extracted by the target model in Fig.~\ref{fig:tsne}. Comparing the DaST with ours, it is obvious that our generated data is widely distributed in the feature space with clearer categorical differences. While, the data generated by DaST have relatively small gap between the data classes and concentrate on the part of the feature space, which is not conducive for the substitute learning and decreases the attack strength. These also further verify the importance of generated data distribution for substitute training.

\begin{figure}
\begin{centering}
\includegraphics[scale=0.37]{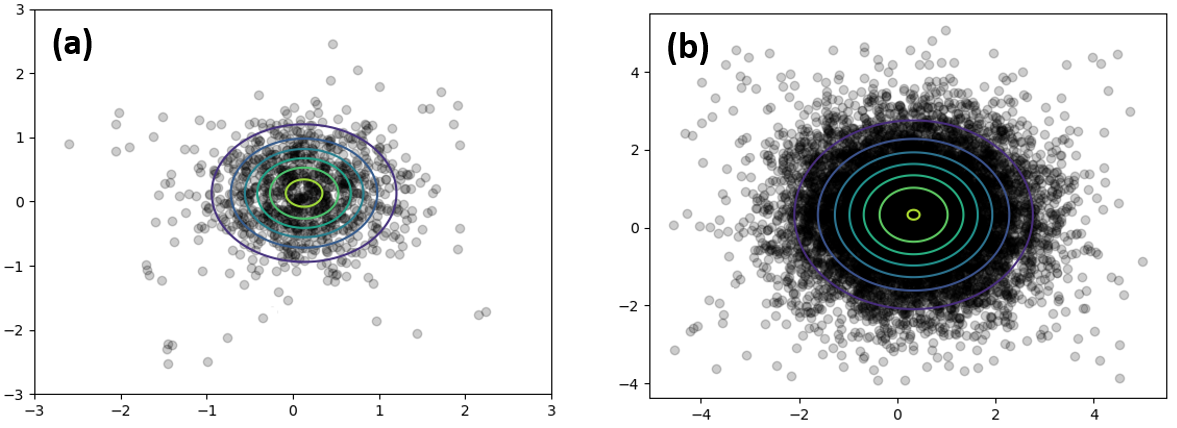}
\par\end{centering}
\caption{The distribution of data in one class on MNIST. (a) Original real data in MNIST. (b) Data generated by our DDG module. 
\label{fig:distribution}}
\end{figure}

\noindent \textbf{Our model can generate diverse data of each class.} 
(1) As shown in Fig.~\ref{fig:generated images}, compared with ours, the DaST generates more similar data in inter-classes, such as the synthetic data are similar in the blue dotted box but various in the red. 
(2) We also visualize the distribution of the same class data collected from MNIST and generated by our method in Fig.~\ref{fig:distribution}. In terms of the amount of data, ours is much larger than MNIST. At the same time, within the class, the data we generate is more widely distributed. These qualitative results demonstrate that our method can generate more diverse data of each class, to further encourage the substitute model to learn from the target.

\begin{figure}
\begin{centering}
\includegraphics[scale=0.37]{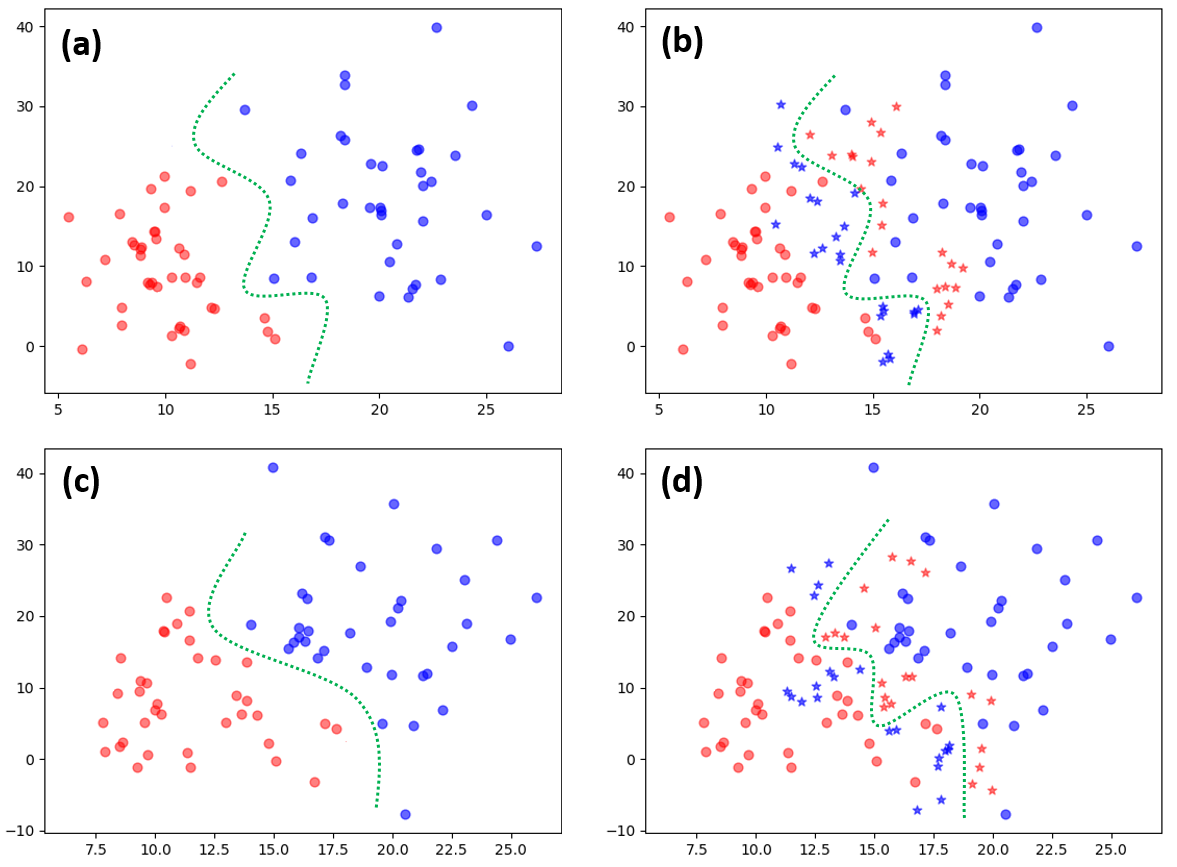}
\par\end{centering}
\caption{Visualization of decision boundaries between two classes (one color indicates one category) in CIFAR-10. The circles represent the normal data, stars mean the adversarial examples, and green dashed lines are the decision boundaries. (a) The real data on target model. (b) The real data and corresponding adversarial examples on target model. (c) The generated data on substitute model. (d) The generated data and corresponding adversarial examples on substitute model. Best viewed in color and zoomed in.
\label{fig:boundary}}
\end{figure}

\noindent \textbf{Our AST strategy can improve the consistency of the decision boundary between the substitute and target models.} As shown in Fig.~\ref{fig:boundary}, the boundaries of the target and substitute models are visualized by input data features using t-SNE. Comparing Fig.~\ref{fig:boundary}(a) and Fig.~\ref{fig:boundary}(b), although they are the same model, the decision boundary is more clear with the adversarial examples around the surface, which demonstrate adversarial examples can help to precisely identify the decision boundary.
Meanwhile, it is clear that, compared to Fig.~\ref{fig:boundary}(b), the visualized decision boundary of substitute model with adversarial examples as in Fig.~\ref{fig:boundary}(c) is more approximated to the target model as shown in Fig.~\ref{fig:boundary}(a), and this intuitively verifies the efficacy of AST to further encourage the mimicking of target `behaviour'.

\section{Conclusion}
This paper focuses on the distribution of generated data for substitute training on black-box attack. 
It proposes a unified substitute model training framework, which contains a diverse data generation module (DDG) and an adversarial substitute training strategy (AST). DDG can generate label-controlled and diverse data to train substitute model.
AST utilizes adversarial examples as boundary data to make the substitute model better fit the decision boundary of the target. Extensive experiments are conducted and results show the method can achieve high attack performance.

\section{Acknowledgement}
This work is supported by NSFC Projects (U62076067), Science and Technology Commission of Shanghai Municipality Projects (19511120700, 19ZR1471800), Shanghai Research and Innovation Functional Program (17DZ2260900), Shanghai Municipal Science and Technology Major Project (2018SHZDZX01) and ZJLab.

{\small
\bibliographystyle{ieee_fullname}
\bibliography{egbib}
}

\end{document}